# Word2Vec and Doc2Vec in Unsupervised Sentiment Analysis of Clinical Discharge Summaries

Qufei Chen
University of Ottawa

qchen037@uottawa.ca

Marina Sokolova
IBDA@Dalhousie University and
University of Ottawa

sokolova@uottawa.ca

## Abstract

In this study, we explored application of Word2Vec and Doc2Vec for sentiment analysis of clinical discharge summaries. We applied unsupervised learning since the data sets did not have sentiment annotations. Note that unsupervised learning is a more realistic scenario than supervised learning which requires an access to a training set of sentiment-annotated data. We aim to detect if there exists any underlying bias towards or against a certain disease. We used SentiWordNet to establish a gold *sentiment* standard for the data sets and evaluate performance of Word2Vec and Doc2Vec methods. We have shown that the Word2vec and Doc2Vec methods complement each other's results in sentiment analysis of the data sets.

## Introduction

Sentiment analysis is the process of identifying opinions expressed in text. Most of the current research of sentiment analysis deals with subjective text, such as customer-written reviews or Twitter [7]. There is considerably less research of sentiment analysis of clinical text in the medical domain. Sentiment analysis in clinical records is quite different from traditional methods. Clinical records are meant be written in an objective way. Most of the written text is descriptive rather than opinionated – it is much more likely to find a sentence that reads "*the patient presented with chest pain*" rather than "*the patient is doing really badly*". Most opinions are not directly stated, but are rather subtle or even unintentional, and is often underlying in the word choice and method of writing of the text. This results in the number of sentiment terms expressed in the text to be rather low (5% - 11% in clinical narratives)(Denecke & Deng, 2015), consequently resulting in a lower accuracy of sentiment analysis (Deng, Declerck, Lendvai, & Denecke, 2016). In addition, the language used in clinical records is very formal and often contains an excess of medical terms and conditions that are not often found elsewhere.

Consequences of detecting opinions in clinical text are also quite significant. It is very important to be able to detect and address bias, especially when dealing with patient care. Identifying the presence of bias can serve to make sure that patients are not discriminated against due to their afflictions, and are all receiving equal quality care. If a bias towards a disease is detected, it can also be indicative of a broader social view towards these diseases and people who are afflicted by them.

Most of the research on sentiment analysis of clinical records has been performed using lexical resources and dictionaries such as SentiWordNet (Esuli & Sebastiani, 2006), WordNetAffect (Strapparava & Valitutti, 2014), and MPQA (Wiebe, Wilson, & Cardie, 2005). While those resources work well for identifying sentiment in highly opinionated text, they do not perform as well on clinical narratives (Deng, Stoehr, & Denecke, Retrieving Attitudes: Sentiment Analysis from Clinical Narratives, 2014). Another challenge for sentiment analysis of clinical records comes from a prohibitively high cost of sentiment annotations of a large set of clinical records. Thus, supervised learning often becomes unfeasible as it is requires an access to a labeled training data. As a result, unsupervised learning from unlabeled data becomes the most realistic scenario in sentiment analysis of clinical text.

The objective of our study is to apply Word2Vec and Doc2Vec in unsupervised sentiment analysis of a set of clinical discharge summaries; the summaries are written by physicians and nurses. We use SentiWordNet to establish a gold *sentiment* standard in a dataset. SentiWordNet is a state-of-the-art sentiment lexicon that has been proven to have a reliable performance compared to other sentiment lexicons (Musto, Semeraro, & Polignano, 2014). We use SentiWordNet to detect the overall sentiment scores and sentiment ratios in our datasets. Using the results as our benchmarks, we then evaluate performance of Word2Vec and Doc2Vec in identifying sentiments in those datasets.

## Related Work

SentiWordNet has proven to be a reliable sentiment Knowledge Base in several sentiment analysis competitions. It obtained the best performing resource when sentiment analysis was conducted on the SemEval2013 dataset (Nakov, Kozareva, Ritter, Rosenthal, Stoyanov, & Wilson, 2013): SentiWordNet's best accuracy was 58.99%, whereas the $2^{nd}$ best accuracy 58.25% was obtained by MPQA. It again obtained the best accuracy on the Stanford Twitter Sentiment dataset (Go, Bhayani, & Huang, 2009): SentiWordNet's performance resulted in 72.42% accuracy, and the $2^{nd}$ best accuracy of 70.75% was again obtained by MPQA (Musto, Semeraro, & Polignano, 2014). Those and other similar results make SentiWordNet our first choice for establishing a gold *sentiment* standard for the data sets.

Deng et al [6] studied how physicians and nurses implicitly and explicitly express their judgments in patient records by using a sentiment lexicon to determine the sentiment of a document. The authors counted the number of positive and negative occurrences in radiology reports and nurse letters, and then compared their results with manual annotations by clinical experts (Deng, Stoehr, & Denecke, Retrieving Attitudes: Sentiment Analysis from Clinical Narratives, 2014). The authors define *sentiment* as information relating the certainty of an observation, information on the health status, or outcome of a medical treatment. However, the accuracy of the proposed method was only 42.0% on nurse letters and 44.6% radiology reports respectively. The authors concluded that a simple method for sentiment analysis is not well suited to analyze sentiment in clinical narratives. We, in other hand, apply advanced Word2Vec and Doc2Vec methods to analyze sentiments in clinical narratives.

Sentiment analysis of tweets under a supervised learning framework has been studied in (Tang, Wei, Yang, Zhou, Liu, & Qin). The authors aimed to learn sentiment specific word embeddings (SSWE), which encodes sentiment information into the word embedding representation. SSWE outperformed traditional neural network models such as Word2Vec on classifying sentiment since traditional models cannot distinguish between words with opposite sentiment that occur in the same context (such as *good* and *bad*). The accuracy of SSWE outperformed Word2Vec by about 10% for both of the lexicons measured (Hu and Liu (Liu & Hu, 2004), MPQA (Wiebe, Wilson, & Cardie, 2005)). Our study, however, differs in two critical ways: i) we work with clinical texts that contain much less sentiment terms than Twitter; ii) we apply unsupervised learning which is a more realistic setting. It is also difficult to obtain access to a large dataset of sentiment labelled clinical data that is required for supervised learning approaches.

## Data Set

The deidentified clinical records used in this research were provided by the i2b2 National Center for Biomedical Computing funded by U54LM008748 and were originally prepared for the Shared Tasks for Challenges in NLP for Clinical Data organized by Dr. Ozlem Uzuner, i2b2 and SUNY.
We were given access to a number of different datasets created by the Informatics for Integrating Biology to the Bedside Center (i2b2).

We have chosen the Obesity NLP Challenge Dataset. This dataset was created by i2b2 for a natural language processing challenge focusing on the extraction of information on obesity and fifteen other common comorbidities (hereinafter referred to as diseases) (Uzuner, 2009). The reason for choosing this dataset is due to the fact that obesity is a highly stigmatized condition in society, with obese individuals often being discriminated against by society (Puhl & Heuer, 2009). It would be important to observe if physicians and nurses exhibit the same bias in their clinical writings.

The dataset consists of 1237 de-identified clinical discharge summaries written by physicians and nurses, as well as a set of annotations that specifies for each discharge summary the occurrence of any number of diseases. The annotations split the occurrence into four classes: Present, Absent, Questionable, and Unknown.

From the given annotations, we were able to separate the discharge summaries into subsets that correspond to each disease. For the purposes of this study, we only used the occurrence class Present as the criteria for determining the separation of subsets. This means that a discharge summary is classified into a particular subset if and only if its class of occurrence for that disease is Present. The number of discharge summaries associated with each disease is listed in Table 1.

As we can see in Table 1, some diseases are not well represented (e.g. *Venous Insufficiency* and *Hypertriglyceridemia*), containing too few discharge summaries with that disease being present. Such datasets are too small to perform an accurate statistical and computational analysis on them. Therefore, for the purposes of this study we have chosen data sets corresponding to three diseases:

1. Hypertension (the most prevalent disease)
2. Diabetes (the second most prevalent)
3. Obesity (known for having a negative stigma and the reason this dataset was chosen)

Each set of discharge summaries relating to one of these diseases will be referred to as a subset.

*Table 1: Frequency Distribution of Diseases*

| Annotated Disease | Number of Summaries |
|---:|---:|
| **Hypertension** | **816** |
| **Diabetes** | **737** |
| Atherosclerotic CV disease (CAD) | 610 |
| Hypercholesterolemia | 459 |
| Heart failure (CHF) | 444 |
| **Obesity** | **443** |
| Gallstones | 178 |
| Osteoarthritis (OA) | 175 |
| GERD | 167 |
| Depression | 158 |
| Obstructive sleep apnea (OSA) | 157 |
| Peripheral vascular disease (PVD) | 147 |
| Asthma | 143 |
| Gout | 123 |
| Hypertriglyceridemia | 25 |
| Venous Insufficiency | 24 |

We performed information extraction with the aid of the Python pandas library, an open source python data processing tool (*https://pandas.pydata.org/*). We were able to extract the discharge summaries corresponding to each dataset by using data frames to cross-reference the discharge summary ID with the IDs in the annotations files. We then sorted the subsets in order of their descending frequency.

Through manual analysis of the distribution of discharge summaries in each subset, we observed that many discharge summaries are associated with more than one disease or condition (e.g. *obesity and diabetes*). Taking into account disease combinations allows a finer-grade sentiment analysis.

Therefore, in addition to the discharge summaries for the three chosen diseases (i.e., hypertension, diabetes, and obesity), we will also be taking their set differences (discharge summaries belonging to *set A* but not *set B*) as well as their intersections (discharge summaries belonging to both *set A* and *set B*) as additional subsets. The resulting twelve subsets along with the number of summaries are shown in Table 2. These are the twelve subsets that will be used for the upcoming experiments in Section Knowledge Base and Methods.

*Table 2: Summary distribution in the subsets*

| Annotated Disease | Number of Summaries |
|---:|---|
| Hypertension | 816 |
| Diabetes | 737 |
| Obesity | 443 |
| Hypertension and Diabetes | 595 |
| Diabetes and not Hypertension | 142 |
| Hypertension and not Diabetes | 221 |
| Obesity and Diabetes | 258 |
| Diabetes and not Obesity | 479 |
| Obesity and not Diabetes | 185 |
| Obesity and Hypertension | 332 |
| Hypertension and not Obesity | 484 |
| Obesity and not Hypertension | 111 |

## Data Preprocessing

The preprocessing of the data was done in Python with the aid of the Natural Language Toolkit (NLTK) (*https://www.nltk.org/*). First, we removed all of the formatting text and special characters from each discharge summary. We then tokenized the remaining discharge summary text. Stop words were also removed using NLTK's English stop words corpus (*https://www.nltk.org/book/ch02.html*). On the next step of preprocessing, we had to decide whether we use stemming or lemmatization to further process the data.

Stemming is the procedure of using a crude heuristic process to chop off the end of a word, for the goal of reducing words in an inflectional or derivational form down to its base form (Manning, 2009). Stemming algorithms usually consists of a static series of steps that are applied to a word to transform it. Lemmatization on the other hand, uses a dictionary and the morphological analysis of words in order to return the base dictionary form of a word (known as a lemma) (Manning, 2009). Lemmas themselves are words that can be found in a dictionary, whereas the result of stemming can and often result in terms that are not standalone English words. For example, when applying a stemming algorithm, the word *see* would just return *s*, while lemmatization would return either *see* or *saw.*

In deciding between stemming and lemmatization, we recalled that our knowledge base SentiWordNet contains sentiments for English dictionary words in WordNet's database (*https://wordnet.princeton.edu/*). Therefore, occurrence of too many inflectional or derivational forms may not be recognized by SentiWordNet. Similarly, occurrence of too many stem terms that are not proper English words is also not beneficial, since these terms will not be recognized by SentiWordNet either.

We tested the result of performing sentiment analysis with SentiWordNet on the datasets with stemming using the Porter Stemming Algorithm (*https://tartarus.org/martin/PorterStemmer/*) and lemmatization separately, and compared them to the original data (with no stemming and no lemmatization) to observe the effects. The results reported in Table 3 show that stemming results in the least number of terms being matched to sentiments in SentiWordNet at an average of only 20.77%, and lemmatization having the best result with an average of 32.04% of terms matched. From the results of this test, we have decided to continue with the lemmatized datasets for the rest of the analysis.

*Table 3: The effects of Stemming and Lemmatization on the percentage of term matches of each dataset in SentiWordNet*

| Dataset | Percentage of Terms Matched in Dataset | | |
|---|---|---|---|
| | Original Data | Stemming | Lemmatization |
| Hypertension | 0.2127 | 0.1553 | 0.2328 |
| Obesity not Hypertension | 0.3605 | 0.1596 | 0.3632 |
| Diabetes not Obesity | 0.2479 | 0.2026 | 0.2756 |
| Obesity not Diabetes | 0.3206 | 0.2321 | 0.4564 |
| Hypertension not Obesity | 0.2474 | 0.2026 | 0.2426 |
| Diabetes | 0.2214 | 0.1909 | 0.3757 |
| Diabetes not Hypertension | 0.3394 | 0.2026 | 0.2794 |
| Obesity | 0.2563 | 0.2536 | 0.2825 |
| Hypertension and Diabetes | 0.2666 | 0.1846 | 0.2763 |
| Obesity and Hypertension | 0.2666 | 0.2780 | 0.2996 |
| Obesity and Diabetes | 0.2666 | 0.2465 | 0.4045 |
| Hypertension not Diabetes | 0.3062 | 0.1837 | 0.3566 |
| **Average Percent Matched** | 0.2760 | 0.2077 | 0.3204 |

After lemmatizing the terms in the datasets, we then proceeded to use WordNet to perform part of speech tagging on each term to prepare them for use with SentiWordNet.

## Knowledge Base and Methods Used

In this study, we will first be using SentiWordNet to determine the overall sentiment scores of each subset, as well as their sentiment ratios which we determine as the proportion of positive terms to negative terms in each dataset. We will then be constructing a Word2Vec model of the data, as well as a Doc2Vec Model of the data, and then compute the cosine similarities of each subset against the other subsets using both the Word2Vec model and the Doc2Vec model.

SentiWordNet

SentiWordNet is a lexical resource used for opinion mining created by Esuli and Sebastiani (2006). It is based on WordNet, a lexical database for the English language that groups words together into sets of synonyms (called synsets). SentiWordNet works on top of that by assigning each synset in WordNet three sentiment scores: a Positive score, a Negative score, and an objective score. The positive score and the negative score make up the subjective portion of the word, and the objective score is calculated by taking one minus the sum of the positive and negative scores (in other words, the objective scores is equal to one minus the total subjective component of the word). For example, the term breakdown contains a positive score of 0, a negative score of 0.25, and an objective score of 0.75. For each of the twelve subsets listed in Table 2, we used SentiWordNet to evaluate the sentiment of each term in the subset, and then calculated the average score for the Positive sentiment, the Negative sentiment, and the Objective sentiment. The resulting sentiment scores are shown in Table 4.

We can see from Table 4 that the majority of each dataset is comprised of objective sentiments (approx. 90% for each dataset), while the positive and negative scores are significantly lower, but with minute differences. This is an expected outcome, since clinical narratives are meant to be written in an objective manner, and therefore less subjective terms should be expected. But to properly compare the scores, we need to analyze if the proportion of subjective (positive and negative) terms are equally distributed between subsets.

*Table 4: Average SentiWordNet sentiment scores of each dataset*

| Data Subset | Positive Score | Negative Score | Objective Score | Overall Sentiment (Negative Score – Positive Score) |
|---|---|---|---|---|
| Hypertension not obesity | 0.0521 | 0.0698 | 0.8781 | 0.0178 |
| Diabetes not obesity | 0.0507 | 0.0708 | 0.8786 | 0.0201 |
| Diabetes not hypertension | 0.0518 | 0.0725 | 0.8757 | 0.0207 |
| diabetes | 0.0510 | 0.0717 | 0.8773 | 0.0208 |
| Hypertension and diabetes | 0.0500 | 0.0710 | 0.8790 | 0.0209 |
| hypertension | 0.0506 | 0.0719 | 0.8775 | 0.0213 |
| Obesity not hypertension | 0.0512 | 0.0728 | 0.8760 | 0.0216 |
| Obesity and diabetes | 0.0507 | 0.0724 | 0.8769 | 0.0217 |
| obesity | 0.0495 | 0.0720 | 0.8785 | 0.0225 |
| Obesity not diabetes | 0.0503 | 0.0736 | 0.8761 | 0.0233 |
| Hypertension not diabetes | 0.0484 | 0.0729 | 0.8787 | 0.0245 |
| Obesity and hypertension | 0.0487 | 0.0747 | 0.8765 | 0.0260 |
| **Overall Average** | 0.0504 | 0.0722 | 0.8774 | 0.0218 |

Figure 1 shows the percentage of positive and negative scores of each subset plotted against each other. We can see from this graph that there are certain datasets that have a higher than average negative score combined with a lower than average positive score (such as the dataset *obesity and hypertension*). This could be indicative of a negative bias in relation to the other subsets. In contrast, the opposite behavior can also be observed – the dataset *hypertension not obesity* contains a lower than average negative sentiment score combined with a higher than average positive sentiment score. This could be indicative of a positive bias in relation to the other data sets.

As we can observe from Figure 1, the most negative overall sentiment occurs when the difference between the negative score and the positive score is high. Similarly, the most positive overall sentiment occurs when that difference is small. Thus, we can deduce that the overall relative sentiment of each subset can be obtained through taking the difference between the negative sentiment and the positive sentiment.

The overall sentiment of each subset is obtained by taking the difference of the negative sentiment score and the positive sentiment score for each subset; we plot them on Figure 2. The overall sentiment scores can also be found in Table 4. We can see from these results that the *obesity and hypertension* dataset has the most negative overall sentiment, with the subset *obesity* being the second most negative, and the subset *hypertension not diabetes* being the third most negative. In contrast, the *hypertension not obesity* subset has the most positive overall sentiment.

*Figure 1: Percentage of negative vs. positive SentiWordNet scores*

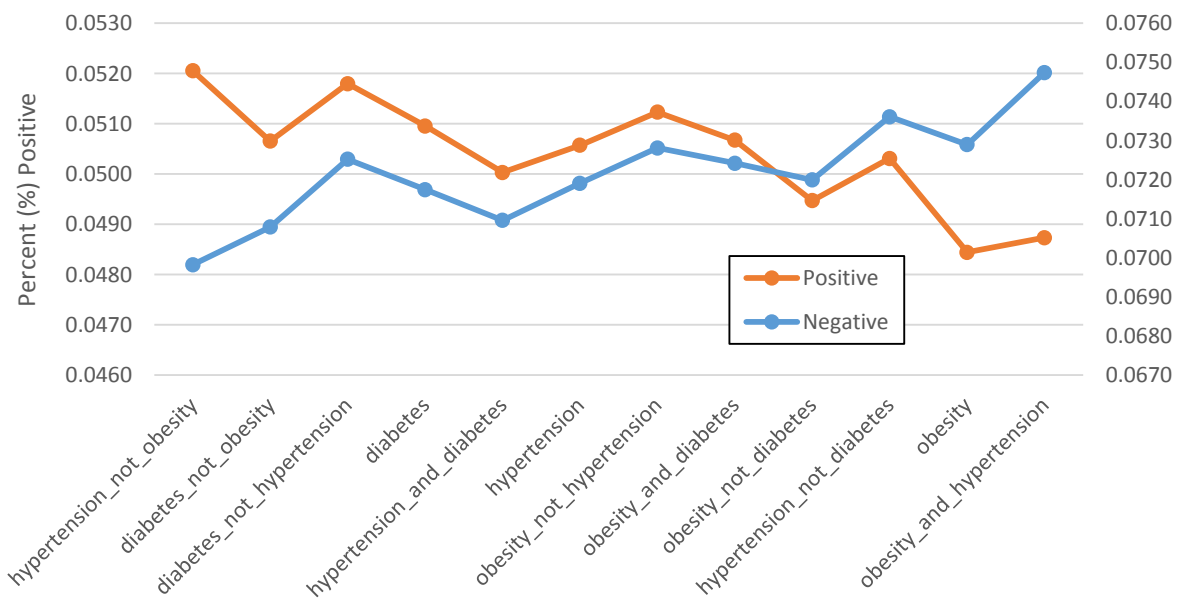

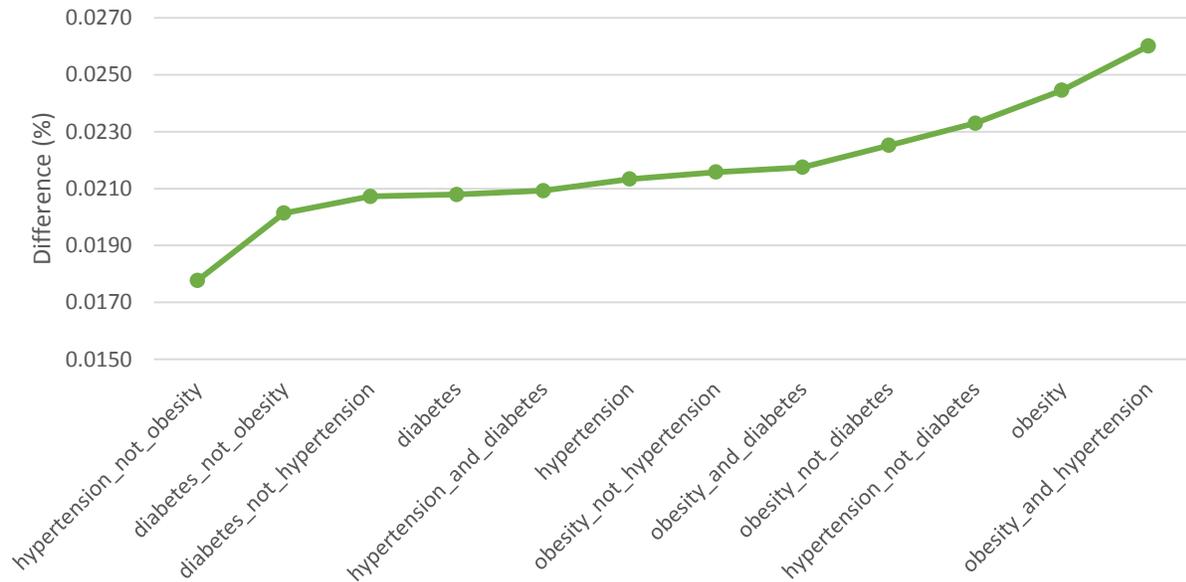

*Figure 2: SentiWordNet overall sentiment (calculated by taking the difference of negative and positive SentiWordNet score of each dataset)*

We also extracted the frequency of words of each sentiment in each subset. From these frequencies, we generated the ratio of positive terms to negative terms of each dataset. We report the raw frequency of positive and negative terms as well as their ratios in Table 5. The subsets are sorted from the smallest ratio of negative terms to the largest ratio. We can see from the data that the positive ratio and negative ratio have an inverse relationship. This relationship is plotted in Figure 3.

From Table 5, we can observe that the subset *obesity not hypertension* contains the highest ratio of positive terms combined with the lowest ratio of negative terms. This indicates that once again this subset has the overall most positive sentiment. Similarly, the subset *obesity and hypertension* contains the highest ratio of negative terms combined with the lowest ratio of positive terms, thus indicating that it possesses the most negative overall sentiment of all the datasets.

The overall sentiment ranking of a few subsets according to the ratio of positive and negative terms differs slightly from their overall SentiWordNet sentiment scores. The subset *obesity not hypertension* has the highest positive ratio and the lowest negative ratio, yet it contains an average overall sentiment score in relation the rest of the datasets. This could indicate that although this dataset contains a large proportion of positive terms, the strengths of those terms are not very high.

*Table 5: Ratio of Positive and negative SentiWordNet terms extracted from each subset*

| Dataset | Number of Positive terms | Number of Negative terms | Ratio Positive (%) | Ratio Negative (%) |
|---|---|---|---|---|
| Obesity not hypertension | 2636 | 2708 | 0.4933 | 0.5067 |
| Hypertension not obesity | 4322 | 4447 | 0.4929 | 0.5071 |
| Diabetes not hypertension | 2931 | 3016 | 0.4929 | 0.5071 |
| Diabetes not obesity | 4321 | 4467 | 0.4917 | 0.5083 |
| diabetes | 4941 | 5114 | 0.4914 | 0.5086 |
| Hypertension and diabetes | 4597 | 4768 | 0.4909 | 0.5091 |
| Obesity not diabetes | 3006 | 3122 | 0.4905 | 0.5095 |
| hypertension | 4971 | 5163 | 0.4905 | 0.5095 |
| Obesity and diabetes | 3503 | 3657 | 0.4892 | 0.5108 |
| Hypertension not diabetes | 3186 | 3338 | 0.4884 | 0.5116 |
| obesity | 4142 | 4341 | 0.4883 | 0.5117 |
| Obesity and hypertension | 3744 | 3948 | 0.4867 | 0.5133 |
| **Overall Average** | 3858.33 | 4007.42 | 0.49 | 0.51 |

*Figure 3: Ratio of Positive and Negative terms of each dataset*

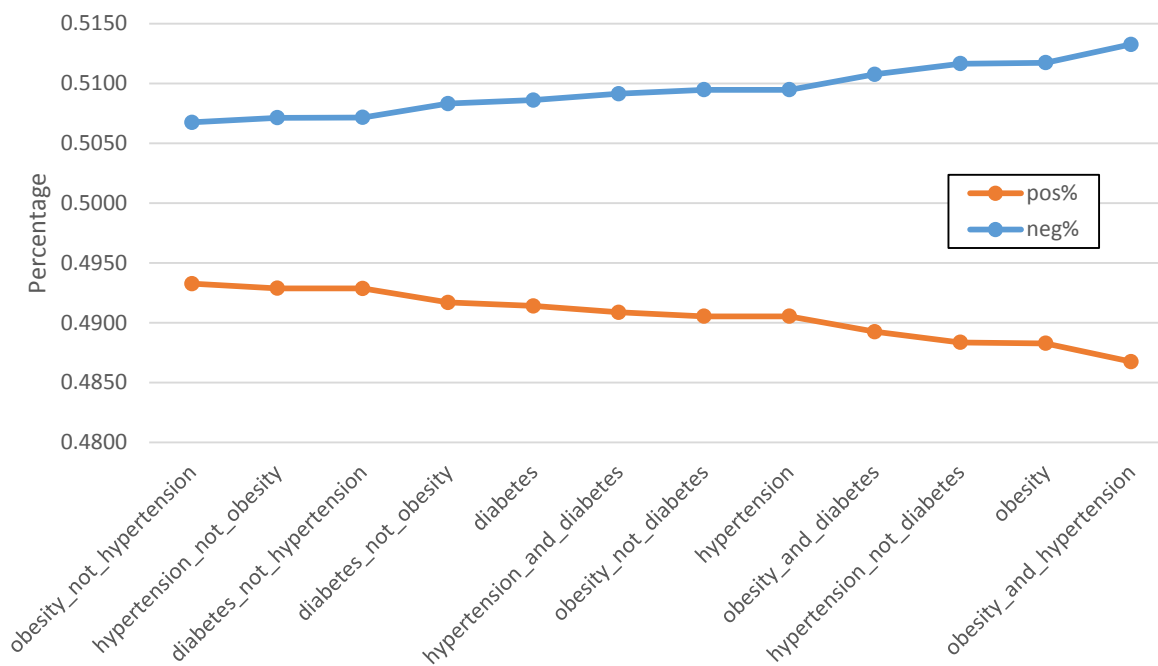

## Word2Vec

Word2Vec is group of unsupervised shallow two-layer neural network models developed by Mikolov et al that produces word embeddings (Mikolov, Chen, Corrado, & Dean, 2013). Word embeddings are the numeric representation of words in the form of vectors. Word2Vec produces word embeddings based on the contextual semantics of words in a text (based on the context that the word occurs in). Words with similar linguistic contexts are mathematically grouped together in a vector space, which preserves the semantic relationship between words. Word2Vec can then use these word embeddings to produce predictions on a word's meaning.

There are two model architectures of Word2Vec that can be used:
- Skip-Gram model
- Continuous Bag-of-Words model (CBOW)

The skip-gram model takes in a word as input and aims to predict a target context, while the continuous bag of words method takes in a context as input and aims to predict a specific word (Deep Learning for Java, 2017). Since we would like to learn the contexts of the words in the data sets, we will be employing the continuous bag-of-words model.

Cosine similarity measures the difference between vectors based on the cosine of the angle between the vectors while normalizing for length (Manning, 2009). This results in a measure for the difference in orientation of the vectors in the vector space rather than a measure of magnitude. We trained a continuous bag-of-words Word2Vec model on the entire dataset (1237 documents) and then proceeded to calculate the cosine similarity between each pair of subsets. The input vector for each subset consisted of all the terms in that subset.

*Table 6: Word2Vec and Doc2Vec corpus sizes*

| Dataset/Model | Corpus Size (Number of Terms) |
|---|---|
| Complete Training Dataset | 804152 |
| Trained Word2Vec Model Vocabulary | 7584 |
| Trained Doc2Vec Model Vocabulary | 12927 |
| Obesity and Diabetes | 213968 |
| Hypertension not Diabetes | 124385 |
| Hypertension | 525672 |
| Diabetes not Hypertension | 91237 |
| Hypertension not Obesity | 311704 |
| Obesity not Hypertension | 62745 |
| Obesity not Diabetes | 101089 |
| Diabetes not Obesity | 316900 |
| Diabetes | 492524 |
| Obesity and Hypertension | 213968 |
| Obesity | 276713 |
| Hypertension and Diabetes | 213968 |

The corpus size of the training dataset and the vocabulary sizes of all the subsets used for evaluation are listed in Table 6. These exact datasets are used for the training and evaluation of both the Word2Vec model and the Doc2Vec model (presented later in this paper). The resulting vocabulary sizes of the trained Word2Vec and Doc2Vec models are also listed in Table 6. Further we will use unique subset ID in lieu of its full descriptive name; the IDs are listed in Table 7.

*Table 7: Dataset IDs*

| Data Subset Name | ID |
|---|---|
| Obesity | 1 |
| Hypertension | 2 |
| Diabetes | 3 |
| Obesity and Hypertension | 4 |
| Obesity not Hypertension | 5 |
| Hypertension not Obesity | 6 |
| Obesity and Diabetes | 7 |
| Obesity not Diabetes | 8 |
| Diabetes not Obesity | 9 |
| Hypertension and Diabetes | 10 |
| Hypertension not Diabetes | 11 |
| Diabetes not Hypertension | 12 |

*Table 8: Word2Vec cosine similarity between datasets*

| ID | 1 | 2 | 3 | 4 | 5 | 6 | 7 | 8 | 9 | 10 | 11 | 12 |
|---|---|---|---|---|---|---|---|---|---|---|---|---|
| 1 | 1.0000 | 0.9987 | 0.9977 | 0.9986 | 0.9831 | 0.9955 | 0.9989 | 0.9966 | 0.9954 | 0.9980 | 0.9947 | 0.9892 |
| 2 | 0.9987 | 1.0000 | 0.9994 | 0.9972 | 0.9820 | 0.9987 | 0.9981 | 0.9945 | 0.9986 | 0.9996 | 0.9953 | 0.9916 |
| 3 | 0.9977 | 0.9994 | 1.0000 | 0.9957 | 0.9828 | 0.9987 | 0.9982 | 0.9913 | 0.9995 | 0.9997 | 0.9922 | 0.9942 |
| 4 | 0.9986 | 0.9972 | 0.9957 | 1.0000 | 0.9722 | 0.9922 | 0.9985 | 0.9936 | 0.9927 | 0.9972 | 0.9912 | 0.9821 |
| 5 | 0.9831 | 0.9820 | 0.9828 | 0.9722 | 1.0000 | 0.9855 | 0.9788 | 0.9855 | 0.9836 | 0.9791 | 0.9855 | 0.9925 |
| 6 | 0.9955 | 0.9987 | 0.9987 | 0.9922 | 0.9855 | 1.0000 | 0.9946 | 0.9919 | 0.9995 | 0.9980 | 0.9948 | 0.9949 |
| 7 | 0.9989 | 0.9981 | 0.9982 | 0.9985 | 0.9788 | 0.9946 | 1.0000 | 0.9917 | 0.9958 | 0.9987 | 0.9899 | 0.9893 |
| 8 | 0.9966 | 0.9945 | 0.9913 | 0.9936 | 0.9855 | 0.9919 | 0.9917 | 1.0000 | 0.9896 | 0.9915 | 0.9982 | 0.9838 |
| 9 | 0.9954 | 0.9986 | 0.9995 | 0.9927 | 0.9836 | 0.9995 | 0.9958 | 0.9896 | 1.0000 | 0.9988 | 0.9920 | 0.9955 |
| 10 | 0.9980 | 0.9996 | 0.9997 | 0.9972 | 0.9791 | 0.9980 | 0.9987 | 0.9915 | 0.9988 | 1.0000 | 0.9920 | 0.9913 |
| 11 | 0.9947 | 0.9953 | 0.9922 | 0.9912 | 0.9855 | 0.9948 | 0.9899 | 0.9982 | 0.9920 | 0.9920 | 1.0000 | 0.9862 |
| 12 | 0.9892 | 0.9916 | 0.9942 | 0.9821 | 0.9925 | 0.9949 | 0.9893 | 0.9838 | 0.9955 | 0.9913 | 0.9862 | 1.0000 |

The results of the cosine similarity between the terms of each subset are shown in Table 8. We can see that all the scores are very similar, with very high values around 0.99. This indicates that the similarity of the terms in each subset to all other subsets is very high. From Table 8, we can observe that the datasets with the highest cosine similarity are subset 10 (*Hypertension and Diabetes*) and subset 3 (*Diabetes*) with a 0.9997 cosine similarity. The subsets with the lowest cosine similarity are subset 5 (*Obesity not Hypertension*) and subset 4 (*Obesity and Hypertension*) with a 0.9722 cosine similarity. In Table 8, the subsets with the highest similarities are highlighted in green while the subsets with the lowest similarities are highlighted in orange.

Doc2Vec

Doc2Vec is an extension of Word2Vec that is applied to a document as a whole instead of individual words. This model was developed by Le and Mikolov, and aims to create a numerical representation of a document rather than a word (Le & Mikolov, 2014). Doc2Vec operates on the logic that the meaning of a word also depends on the document that it occurs in. The vectors generated by Doc2Vec can be used for finding similarities between documents. We trained a Doc2Vec model on all of the discharge summaries in the complete dataset (1237 discharge summaries), and then proceeded to convert each subset into its word embedding representation (also known as an inferred vector). We then computed the cosine similarities between each pair of subsets. The results of the cosine similarities scores between subsets are shown in Table 9.

*Table 9: Doc2Vec cosine similarity scores between datasets*

| ID | 1 | 2 | 3 | 4 | 5 | 6 | 7 | 8 | 9 | 10 | 11 | 12 |
|---|---|---|---|---|---|---|---|---|---|---|---|---|
| 1 | 1.0000 | 0.7943 | 0.8116 | 0.9229 | 0.6842 | 0.5638 | 0.8976 | 0.7973 | 0.6670 | 0.7750 | 0.6681 | 0.5548 |
| 2 | 0.7943 | 1.0000 | 0.8921 | 0.7697 | 0.7349 | 0.7927 | 0.7781 | 0.7538 | 0.8122 | 0.8950 | 0.6308 | 0.6643 |
| 3 | 0.8116 | 0.8921 | 1.0000 | 0.8280 | 0.6241 | 0.7233 | 0.8248 | 0.6292 | 0.7866 | 0.9623 | 0.5121 | 0.6806 |
| 4 | 0.9229 | 0.7697 | 0.8280 | 1.0000 | 0.6146 | 0.5868 | 0.9430 | 0.7833 | 0.7031 | 0.8026 | 0.6045 | 0.5341 |
| 5 | 0.6842 | 0.7349 | 0.6241 | 0.6146 | 1.0000 | 0.6180 | 0.6774 | 0.6585 | 0.6641 | 0.6213 | 0.7058 | 0.7288 |
| 6 | 0.5638 | 0.7927 | 0.7233 | 0.5868 | 0.6180 | 1.0000 | 0.6335 | 0.5787 | 0.9020 | 0.7281 | 0.6502 | 0.6470 |
| 7 | 0.8976 | 0.7781 | 0.8248 | 0.9430 | 0.6774 | 0.6335 | 1.0000 | 0.7484 | 0.7191 | 0.7904 | 0.6660 | 0.6074 |
| 8 | 0.7973 | 0.7538 | 0.6292 | 0.7833 | 0.6585 | 0.5787 | 0.7484 | 1.0000 | 0.6062 | 0.6724 | 0.7722 | 0.3546 |
| 9 | 0.6670 | 0.8122 | 0.7866 | 0.7031 | 0.6641 | 0.9020 | 0.7191 | 0.6062 | 1.0000 | 0.7842 | 0.6192 | 0.7292 |
| 10 | 0.7750 | 0.8950 | 0.9623 | 0.8026 | 0.6213 | 0.7281 | 0.7904 | 0.6724 | 0.7842 | 1.0000 | 0.5520 | 0.6708 |
| 11 | 0.6681 | 0.6308 | 0.5121 | 0.6045 | 0.7058 | 0.6502 | 0.6660 | 0.7722 | 0.6192 | 0.5520 | 1.0000 | 0.4814 |
| 12 | 0.5548 | 0.6643 | 0.6806 | 0.5341 | 0.7288 | 0.6470 | 0.6074 | 0.3546 | 0.7292 | 0.6708 | 0.4814 | 1.0000 |

From Table 9, we can observe that the most similar pair of subsets are subset 10 (*Hypertension and Diabetes*) and subset 3 (*Diabetes*) with a cosine similarity score of 0.9623. The least similar pair of subsets are subset 12 (*Diabetes not Hypertension*) and subset 8 (*Obesity not Diabetes*) with a cosine similarity score of 0.3546. In Table 9, the subsets with the highest similarities are highlighted in green while the subsets with the lowest similarities are highlighted in orange.

## Analysis of Results

We performed a statistical analysis on the SentiWordNet sentiment scores in order to evaluate whether or not the results were statistically significant. We performed an independent two-sample Welch's t-test on the SentiWordNet sentiment scores between each pair of subsets. The Welch's t-test is a variation of the Student's t-test. It performs better than the Student's t-test when the samples being compared have unequal variances and sample sizes, and performs the same as the Student's t-test when the variances and sample sizes are equal (Lakens, 2015). Since all our subsets are of different sizes, and we cannot assume equal variance between subsets, we performed an independent two-sample Welch's t-test on each pair of subsets for the positive sentiment score, the negative sentiment score, and the overall sentiment score.

Our null hypothesis for each test is that the means of both samples are equal ($\mu_1 = \mu_2$). The p-value of each test represents the probability of finding the observed results when the null hypothesis is true – in other words, it represents the smallest level of significance at which the null hypothesis can be rejected (Investopedia). We used the Welch's t-test function in the statistics module of SciPy, an open source python library for scientific computing (www.scipy.org), to calculate the p-values.

The results of the Welch's t-test are shown in Table 10, Table 11, and Table 12.

*Table 10: Welch's t-test p-value results for SentiWordNet positive sentiment*

| ID | 1 | 1 | 3 | 4 | 5 | 6 | 7 | 8 | 9 | 10 | 11 | 12 |
|---|---|---|---|---|---|---|---|---|---|---|---|---|
| 1 | 1.000 | 0.307 | 0.291 | 0.952 | 0.581 | 0.553 | 0.837 | 0.850 | 0.457 | 0.798 | 0.876 | 0.365 |
| 2 | 0.307 | 1.000 | 0.967 | 0.294 | 0.746 | 0.689 | 0.449 | 0.465 | 0.804 | 0.441 | 0.436 | 0.988 |
| 3 | 0.291 | 0.967 | 1.000 | 0.279 | 0.721 | 0.662 | 0.429 | 0.446 | 0.775 | 0.421 | 0.417 | 0.984 |
| 4 | 0.952 | 0.294 | 0.279 | 1.000 | 0.554 | 0.526 | 0.796 | 0.810 | 0.435 | 0.757 | 0.835 | 0.349 |
| 5 | 0.581 | 0.746 | 0.721 | 0.554 | 1.000 | 0.978 | 0.727 | 0.731 | 0.918 | 0.738 | 0.704 | 0.761 |
| 6 | 0.553 | 0.689 | 0.662 | 0.526 | 0.978 | 1.000 | 0.719 | 0.725 | 0.882 | 0.729 | 0.694 | 0.715 |
| 7 | 0.837 | 0.449 | 0.429 | 0.796 | 0.727 | 0.719 | 1.000 | 0.994 | 0.615 | 0.972 | 0.967 | 0.495 |
| 8 | 0.850 | 0.465 | 0.446 | 0.810 | 0.731 | 0.725 | 0.994 | 1.000 | 0.625 | 0.967 | 0.974 | 0.506 |
| 9 | 0.457 | 0.804 | 0.775 | 0.435 | 0.918 | 0.882 | 0.615 | 0.625 | 1.000 | 0.618 | 0.594 | 0.817 |
| 10 | 0.798 | 0.441 | 0.421 | 0.757 | 0.738 | 0.729 | 0.972 | 0.967 | 0.618 | 1.000 | 0.938 | 0.493 |
| 11 | 0.876 | 0.436 | 0.417 | 0.835 | 0.704 | 0.694 | 0.967 | 0.974 | 0.594 | 0.938 | 1.000 | 0.480 |
| 12 | 0.365 | 0.988 | 0.984 | 0.349 | 0.761 | 0.715 | 0.495 | 0.506 | 0.817 | 0.493 | 0.480 | 1.000 |

*Table 11: : Welch's t-test p-value results for SentiWordNet negative sentiment*

| ID | 1 | 1 | 3 | 4 | 5 | 6 | 7 | 8 | 9 | 10 | 11 | 12 |
|---|---|---|---|---|---|---|---|---|---|---|---|---|
| 1 | 1.000 | 0.771 | 0.736 | 0.614 | 0.983 | 0.376 | 0.897 | 0.817 | 0.546 | 0.574 | 0.852 | 0.924 |
| 2 | 0.771 | 1.000 | 0.960 | 0.419 | 0.818 | 0.530 | 0.885 | 0.982 | 0.738 | 0.773 | 0.644 | 0.870 |
| 3 | 0.736 | 0.960 | 1.000 | 0.395 | 0.787 | 0.564 | 0.849 | 0.948 | 0.775 | 0.811 | 0.614 | 0.836 |
| 4 | 0.614 | 0.419 | 0.395 | 1.000 | 0.642 | 0.170 | 0.540 | 0.492 | 0.273 | 0.288 | 0.773 | 0.578 |
| 5 | 0.983 | 0.818 | 0.787 | 0.642 | 1.000 | 0.455 | 0.926 | 0.851 | 0.614 | 0.641 | 0.853 | 0.947 |
| 6 | 0.376 | 0.530 | 0.564 | 0.170 | 0.455 | 1.000 | 0.472 | 0.571 | 0.777 | 0.737 | 0.314 | 0.480 |
| 7 | 0.897 | 0.885 | 0.849 | 0.540 | 0.926 | 0.472 | 1.000 | 0.915 | 0.653 | 0.683 | 0.764 | 0.980 |
| 8 | 0.817 | 0.982 | 0.948 | 0.492 | 0.851 | 0.571 | 0.915 | 1.000 | 0.754 | 0.786 | 0.697 | 0.900 |
| 9 | 0.546 | 0.738 | 0.775 | 0.273 | 0.614 | 0.777 | 0.653 | 0.754 | 1.000 | 0.961 | 0.455 | 0.651 |
| 10 | 0.574 | 0.773 | 0.811 | 0.288 | 0.641 | 0.737 | 0.683 | 0.786 | 0.961 | 1.000 | 0.477 | 0.680 |
| 11 | 0.852 | 0.644 | 0.614 | 0.773 | 0.853 | 0.314 | 0.764 | 0.697 | 0.455 | 0.477 | 1.000 | 0.793 |
| 12 | 0.924 | 0.870 | 0.836 | 0.578 | 0.947 | 0.480 | 0.980 | 0.900 | 0.651 | 0.680 | 0.793 | 1.000 |

*Table 12: Welch's t-test p-value results for SentiWordNet overall sentiment*

| ID | 1 | 1 | 3 | 4 | 5 | 6 | 7 | 8 | 9 | 10 | 11 | 12 |
|---|---|---|---|---|---|---|---|---|---|---|---|---|
| 1 | 1.000 | 0.450 | 0.377 | 0.725 | 0.561 | 0.116 | 0.547 | 0.682 | 0.309 | 0.399 | 0.802 | 0.431 |
| 2 | 0.450 | 1.000 | 0.892 | 0.274 | 0.960 | 0.384 | 0.924 | 0.796 | 0.770 | 0.919 | 0.661 | 0.895 |
| 3 | 0.377 | 0.892 | 1.000 | 0.224 | 0.871 | 0.463 | 0.827 | 0.707 | 0.873 | 0.974 | 0.578 | 0.989 |
| 4 | 0.725 | 0.274 | 0.224 | 1.000 | 0.382 | 0.061 | 0.357 | 0.472 | 0.181 | 0.240 | 0.568 | 0.277 |
| 5 | 0.561 | 0.960 | 0.871 | 0.382 | 1.000 | 0.439 | 0.974 | 0.860 | 0.769 | 0.893 | 0.743 | 0.873 |
| 6 | 0.116 | 0.384 | 0.463 | 0.061 | 0.439 | 1.000 | 0.373 | 0.312 | 0.575 | 0.448 | 0.229 | 0.530 |
| 7 | 0.547 | 0.924 | 0.827 | 0.357 | 0.974 | 0.373 | 1.000 | 0.875 | 0.718 | 0.852 | 0.748 | 0.836 |
| 8 | 0.682 | 0.796 | 0.707 | 0.472 | 0.860 | 0.312 | 0.875 | 1.000 | 0.611 | 0.731 | 0.877 | 0.727 |
| 9 | 0.309 | 0.770 | 0.873 | 0.181 | 0.769 | 0.575 | 0.718 | 0.611 | 1.000 | 0.849 | 0.491 | 0.900 |
| 10 | 0.399 | 0.919 | 0.974 | 0.240 | 0.893 | 0.448 | 0.852 | 0.731 | 0.849 | 1.000 | 0.601 | 0.966 |
| 11 | 0.802 | 0.661 | 0.578 | 0.568 | 0.743 | 0.229 | 0.748 | 0.877 | 0.491 | 0.601 | 1.000 | 0.610 |
| 12 | 0.431 | 0.895 | 0.989 | 0.277 | 0.873 | 0.530 | 0.836 | 0.727 | 0.900 | 0.966 | 0.610 | 1.000 |

The lowest p-value for each test is highlighted in orange. Table 10 lists the p-values of the Welch's t-test for the SentiWordNet positive sentiment scores. The lowest p-value observed is 0.291 between subset 1 (*Obesity*) and subset 3 (*Diabetes*). This is a very high p-value, indicating that 29.1% of the time we would accept the null hypothesis. Or in other words, there is only a 70.9% chance that the difference is statistically significant.

Table 11 lists the p-values of the Welch's t-test for the SentiWordNet negative sentiment scores. The lowest p-value observed is 0.170 between subset 4 (*Obesity and Hypertension*) and subset 6

(*Hypertension not Obesity*). Although this p-value is better than the p-value for the positive sentiment score, there is still only an 83.0% chance that the difference is statistically significant.

Table 12 lists the p-values of the Welch's t-test for the SentiWordNet overall sentiment scores. We can observe that the lowest p-value is 0.061, and occurs again between subset 4 (*Obesity and Hypertension*) and subset 6 (*Hypertension not Obesity*). This p-value indicates that there is a 93.9% chance that the difference is statistically significant. Although this value is not quite 95% (the accepted standard), it is still considered quite high since it is below 0.1.

Recall that from Figure 2 we can discern that the subset *Obesity and Hypertension* possesses the most negative overall sentiment, while the subset *Hypertension not Obesity* possesses the most positive overall sentiment. Through conducting a Welch's t-test, we have determined that the difference between these two datasets have a 93.9% chance of being statistically significant.

## Discussion of Sentiments

We aim to empirically analyze the sentiment agreement (Cagan, Frank, & Tsarfaty, 2017) between Word2Vec, Doc2Vec, and SentiWordNet. We examine the most similar and the most different *in sentiment* pairs of subsets produced by each of these methods.

**Most Similar Subsets.** When determining the most similar *in sentiment* pair of subsets from the data, the Doc2Vec cosine similarity score, Word2Vec cosine similarity score, the SentiWordNet sentiment score, and SentiWordNet sentiment ratios all show the same result: the most similar pair of subsets are subset 10 (*Hypertension and Diabetes*) and subset 3 (*Diabetes*). The characteristics of these two subsets are summarized in Table 13.

*Table 13: Most Similar subsets*

| | Most Similar Subsets | |
|---:|---|---|
| Dataset | Hypertension and Diabetes (ID = 10) | Diabetes (ID = 3) |
| Doc2Vec Score | 0.9623 | |
| Word2Vec Score | 0.9997 | |
| SentiWordNet Positive Score | 0.0500 | 0.510 |
| SentiWordNet Negative Score | 0.0710 | 0.0717 |
| SentiWordNet Overall Score | 0.0209 | 0.0208 |
| Welch's t-test positive p-value | 0.421 | |
| Welch's t-test negative p-value | 0.811 | |
| Welch's t-test overall p-value | 0.974 | |
| SentiWordNet positive terms | 4597 | 4941 |
| SentiWordNet negative terms | 4768 | 5114 |
| SentiWordNet positive terms ratio | 0.4909 | 0.4914 |
| SentiWordNet negative terms ratio | 0.5091 | 0.5086 |

We can observe from the Doc2Vec cosine similarity results (shown in Table 9) that the most similar pair of subset 10 (*Hypertension and Diabetes*) and subset 3 (*Diabetes*) has a cosine similarity score of 0.9623. The corresponding Word2Vec cosine similarity scores (shown in Table 8) for these two datasets is 0.9997, which is also the highest Word2Vec cosine similarity score observed out of all pairs of subsets. This indicates that these two datasets are the two most similar subsets according to the Doc2Vec model and Word2Vec model.

As shown in Table 13, the SentiWordNet sentiment scores for these two subsets are very similar (the subset *Hypertension and Diabetes* having an overall sentiment of 0.0209 and the subset *Diabetes* having an overall sentiment score of 0.0208). In Figure 2 we can actually see that the subset *Hypertension and Diabetes* and the subset *Diabetes* lie right next to each other on the plot, with the subset *Diabetes* having the fourth most positive sentiment, and the subset *Hypertension and Diabetes* having the fifth most positive sentiment. The difference between their overall sentiment scores is only 0.001. In fact, these two subsets are tied for the most similar overall sentiment (the pair of subsets *Diabetes* and *Diabetes and not Hypertension* also have the same overall sentiment difference of 0.0001).

When examining the results of a Welch's t-test on the sentiment scores of these two datasets, we can observe that the p-value for the positive sentiment score is 0.421, the p-value for the negative sentiment score is 0.811, and the p-value for the overall sentiment score is 0.974. These values are all very high, which indicates that the null hypothesis of equal means should be accepted. Accepting the null hypothesis indicates that there is no statistically significant difference between the two datasets. This result supports our observation that these two datasets have a very high similarity in sentiment.

The ratio of positive and negative sentiment terms between the two subsets are also very similar (with the subset *Hypertension and* Diabetes having a positive terms ratio of 0.4900, and a negative terms ratio of 0.5091*,* and the subset Diabetes having a positive terms ratio of 0.4914, and a negative terms ratio of 0.5086). As shown in Figure 3, the subset *Diabetes* has the fifth smallest ratio of negative terms, as well as the fifth smallest ratio of positive terms. Very similarly, the subset *Hypertension and Diabetes* has the sixth smallest ratio of negative terms, as well as the sixth smallest ratio of positive terms. We can observe that the distributions of positive and negative words of the two datasets are very similar.

For determining the most similar *in sentiment* pair of subsets, the Doc2Vec cosine similarity, the Word2Vec cosine similarity, the SentiWordNet sentiment scores, and the SentiWordNet sentiment ratios seem to all agree that the subset *Hypertension and Diabetes* and the subset *Diabetes* are the most similar in sentiment.

**Most Different Subsets.** When determining the most different *in sentiment* pair of subsets from the data, the Doc2Vec cosine similarity score, Word2Vec cosine similarity score, the SentiWordNet sentiment score, and SentiWordNet sentiment ratios all show different results. The pair of subsets determined to be most difference by each evaluation method are:

- **Doc2Vev**: subset 8 (*Obesity not Diabetes)* and subset 12 (*Diabetes not Hypertension*)
- **SentiWordNet Sentiment Score**: subset 6 (*Hypertension and not Obesity*) and subset 4 (*Obesity and Hypertension*
- **Word2Vec** and **SentiWordNet Sentiment Ratio**: subset 5 (*Obesity and not Hypertension*) and subset 4 (*Obesity and Hypertension)*

**Doc2Vec Result.** According to the Doc2Vec cosine similarity results (shown in Table 9), the most different *in sentiment* pair of subsets are subset 8 (*Obesity not Diabetes* ) and subset 12 (*Diabetes not Hypertension*) at a cosine similarity score of 0.3546. The characteristics of these two subsets are summarized in Table 14.

The Word2Vec cosine similarity of these two datasets is 0.9838, which is a relatively average similarity score in comparison to the rest of the datasets (the highest similarity being 0.9997 and the lowest similarity being 0.9722). This result is not very informative.
When examining the SentiWordNet scores of these two subsets, we can observe that the subset *Diabetes and not Hypertension* is the third most positive overall, and the subset *Obesity and not Diabetes* is the ninth most positive overall (see Figure 2). While there is a relatively large difference of 0.0026 between their overall sentiment scores, this is not the largest difference in overall sentiment that exists. The results of the Welch's t-test on these two subsets also strongly indicate that the difference between the two datasets are not significant (the overall p-value of 0.727 signifies that there is only a 27.3% chance that the difference is statistically significant).

*Table 14: Most different subsets according to Doc2Vec*

|  | **Most Different Subsets** | |
| ---: | --- | --- |
| Dataset | *Obesity not Diabetes* (ID = 8) | *Diabetes not Hypertension* (ID = 12) |
| Doc2Vec Score | 0.3546 | |
| Word2Vec Score | 0.9838 | |
| SentiWordNet Positive Score | 0.0503 | 0.0518 |
| SentiWordNet Negative Score | 0.0736 | 0.0725 |
| SentiWordNet Negative - Positive | 0.0233 | 0.0207 |
| Welch's t-test positive p-value | 0.506 | |
| Welch's t-test negative p-value | 0.900 | |
| Welch's t-test overall p-value | 0.727 | |
| SentiWordNet positive terms | 3006 | 2931 |
| SentiWordNet negative terms | 3122 | 3016 |
| SentiWordNet positive terms ratio | 0.4905 | 0.4929 |
| SentiWordNet negative terms ratio | 0.5095 | 0.5071 |

Examining the ratio of sentiment terms in each subset gives a comparable result – the subset *Diabetes and not Hypertension* has the third lowest ratio of negative terms (0.5071), while the subset *Obesity and not Diabetes* has the seventh lowest ratio of negative terms (0.5095). Although there exists a difference of 0.0024, the difference is not the largest out of all the subsets - the largest difference being 0.0066 occurring between the subsets *Obesity and Hypertension* and *Obesity not Hypertension*.

**SentiWordNet Sentiment Score Result.** According to the SentiWordNet overall sentiment results (shown in Table 4), the most different pair of datasets are subset 6 (*Hypertension and not Obesity* ) and subset 4 (*Obesity and Hypertension*). The difference between their overall sentiment scores is 0.0082, the largest difference of any two datasets. The characteristics of these two subsets are summarized in Table 15.

When observing the SentiWordNet overall sentiment scores of these two subsets (shown in Table 4), we can observe that the subset *Hypertension and not Obesity* is the most positive subset overall, while the subset *Obesity and Hypertension* is the most negative overall (see Figure 2). These two subsets have the largest difference of positive sentiment scores and negative sentiment scores. Observing the results of the Welch's t-test on these two subsets shows a similar result – the p-value for the overall sentiment is 0.061, the smallest and most significant score of all the datasets. This signifies that we can reject the null hypothesis with a 93.9% confidence, i.e., there is a 93.9% chance that difference between these two datasets is statistically significant.

*Table 15: Most different subsets according to SentiWordNet overall sentiment*

|  | Most Different Subsets | |
| --- | --- | --- |
| Dataset | Hypertension and not Obesity (ID = 6) | Obesity and Hypertension (ID = 4) |
| Doc2Vec Score | 0.5868 | |
| Word2Vec Score | 0.9922 | |
| SentiWordNet Positive Score | 0.0521 | 0.0487 |
| SentiWordNet Negative Score | 0.0698 | 0.0747 |
| SentiWordNet Negative - Positive | 0.0178 | 0.0260 |
| Welch's t-test positive p-value | 0.526 | |
| Welch's t-test negative p-value | 0.170 | |
| Welch's t-test overall p-value | 0.061 | |
| SentiWordNet positive terms | 4322 | 3744 |
| SentiWordNet negative terms | 4447 | 3948 |
| SentiWordNet positive terms ratio | 0.4929 | 0.4867 |
| SentiWordNet negative terms ratio | 0.5071 | 0.5133 |

Examining the ratio of sentiment terms in each subset gives a similar result – the subset *Hypertension and not Obesity* has the second highest ratio of positive terms combined with the second lowest ratio of negative terms (making it second most positive overall). The subset *Obesity and Hypertension* has the lowest ratio of positive terms combined with the highest ratio of negative terms (making it the most negative overall) (see Table 5). In terms of the difference in ratio of positive and negative terms, these two datasets are the second most different pair of datasets (the most different pair of subsets being *Obesity and not Hypertensio*n and *Obesity and Hypertension*), indicating that they are indeed quite different in sentiment.

The Word2Vec cosine similarity of these two subsets is 0.9922, and is a relatively average similarity score in comparison to the rest of the datasets. The Doc2Vec cosine similarity of these two subsets in 0.5868, and is also a relatively average similarity score in comparison to the rest of the datasets. Neither of these cosine similarity scores shows the same differences as the SentiWordNet sentiment scores.

**Word2Vec and SentiWordNet Sentiment Ratio Result**. According to the Word2Vec cosine similarity results (shown in Table 9), the most different pair of datasets are subset 5 (*Obesity and not Hypertension*) and subset 4 (*Obesity and Hypertension)* at a cosine similarity score of 0.9722. The Doc2Vec cosine similarity of these two datasets is 0.6146, which is a relatively average similarity score in comparison to the rest of the datasets (the highest similarity being 0.9623 and the lowest similarity being 0.3546). This result does not reflect the same differences as the Word2Vec cosine similarity score.

When observing the SentiWordNet overall sentiment scores of these two subsets (shown in Table 4), we can see that subset 5 (*Obesity and not Hypertension*) is the seventh most positive overall and subset 4 (*Obesity and Hypertension*) is the twelfth most positive (or the most negative) (see Figure 2). While there does exist a difference between these subsets, this difference is not as large as some other pairs of subsets (such as *Hypertensions and not Obesity* and *Obesity and Hypertension*). Additionally, when analyzing the result of the Welch's t-test for these two subsets, the p-value for all three tests is very high, which strongly indicates that the difference between the two datasets is not statistically significant (an overall sentiment p-value of 0.382 signifies that there is only a 67.8% change that the difference is statistically significant).

When observing the ratio of sentiment terms in each subset, we can see that subset 5 (*Obesity and not Hypertension*) contains the highest ratio of positive terms (0.4933) combined with the lowest ratio of negative terms (0.5067), and subset 4 (*Obesity and Hypertension*) contains the lowest ratio of positive terms (0.4867) combined with the highest ration of negative terms (0.5133). In terms of the difference in ratio of positive and negative terms, these two datasets are indeed the pair of subsets that are most different in sentiment. The ratio of sentiment terms supports the Word2Vec result that these two subsets are the most different.

*Table 16: Most different subsets according to Word2Vec and SentiWordNet sentiment ratio*

| Dataset | Most Different Subsets | |
|---|---|---|
| | Obesity and not Hypertension (ID = 5) | Obesity and Hypertension (ID = 4) |
| Doc2Vec Score | 0.6146 | |
| Word2Vec Score | 0.9722 | |
| SentiWordNet Positive Score | 0.0512 | 0.0487 |
| SentiWordNet Negative Score | 0.0728 | 0.0747 |
| SentiWordNet Negative - Positive | 0.0216 | 0.0260 |
| Welch's t-test positive p-value | 0.554 | |
| Welch's t-test negative p-value | 0.642 | |
| Welch's t-test overall p-value | 0.382 | |
| SentiWordNet positive terms | 2636 | 3744 |
| SentiWordNet negative terms | 2708 | 3948 |
| SentiWordNet positive terms ratio | 0.4933 | 0.4867 |
| SentiWordNet negative terms ratio | 0.5067 | 0.5133 |

**Overall Results for Most Different *In Sentiment* Subsets.** The pair of documents determined by Doc2Vec cosine similarity to be most different in sentiment are subset 8 (*Obesity not Diabetes*) and subset 12 (*Diabetes not Hypertension*) (see Table 14). This result is not supported by the results from the Word2Vec cosine similarity, the SentiWordNet sentiment score, nor the SentiWordNet sentiment ratio.

The pair of documents determined by the overall SentiWordNet sentiment score to be the most different in sentiment are subset 6 (*Hypertension and not Obesity*) and subset 4 (*Obesity and Hypertension*) (see Table 15). This result is supported by the result of the Welch's t-test (having a 93.9% chance that the difference between these two subsets is statistically significant). It is also to an extent supported by the SentiWordNet sentiment ratios as being the second most different pair of subsets. But there is no strong evidence from Word2Vec cosine similarity nor Doc2Vec cosine similarity between the subsets that supports this result.

The Word2Vec cosine similarity and the SentiWordNet sentiment ratios both determined subset 5 (*Obesity and not Hypertension*) and subset 4 (*Obesity and Hypertension*) to be the most different in sentiment (see Table 16). But this result is not strongly supported by either Doc2Vec cosine similarity or by SentiWordNet sentiment scores.

## Conclusions and Future Work

In this study, we explored the application of the unsupervised machine learning models Word2Vec and Doc2Vec on detecting sentiments of clinical discharge summaries. We have discovered that Word2vec and Doc2Vec have the same performance as SentiWordNet on the task of predicting datasets with similar sentiments. We have also discovered that Doc2Vec does not perform well in predicting datasets with different sentiments. Word2Vec on the other hand, is able to distinguish between subsets with a different ratios of sentiment terms (positive and negative), but not by their overall sentiment score. We used Welch's t-test to evaluate statistical significance of the obtained results.

We performed sentiment analysis on 12 subsets of clinical discharge summaries to evaluate the overall sentiment scores of subsets relating to certain diseases. We found through using the traditional sentiment lexicon SentiWordNet that the subset *Hypertension not Obesity* contained the most positive overall sentiment relative to the other subsets, and the subset *Obesity and Hypertension* contained the most negative overall sentiment relative to the other subsets. On the two datasets, Welch's t-test resulted in a p-value of 0.061, indicating that there is 93.9% chance that the difference between the subsets are statistically significant.

Future work in this area include extending the scope of the analysis to different datasets, such as age or gender, and including different lexical resources (such as MPQA (Wiebe, Wilson, & Cardie, 2005) and Bing Liu's Opinion Lexicon (Liu & Hu, 2004)) as part of our analysis. Furthermore, we plan develop a method to evaluate the performance of unsupervised machine learning models (such as Word2Vec and Doc2Vec) in determining sentiment of a document.